\documentclass[fleqn,10pt,twocolumn]{ICCAS2019}

\usepackage{graphicx}
\usepackage{caption}
\usepackage{subcaption}
\usepackage{amsmath}
\usepackage{hhline}
\usepackage{dblfloatfix}
\usepackage{url}
\usepackage{tikz}
\newcommand*\circled[1]{\tikz[baseline=(char.base)]{
		\node[shape=circle,draw,inner sep=0.5pt] (char) {#1};}}
\begin{document}
	
	\title{A Dual Memory Structure for Efficient Use of Replay Memory in\\ Deep Reinforcement Learning}
	
	\author{Wonshick Ko${}^{1}$ and Dong Eui Chang${}^{1*}$ }
	
	\affils{ ${}^{1}$School of Electrical Engineering, KAIST, \\
		Daejeon, 34141, Korea (kows93@kaist.ac.kr, dechang@kaist.ac.kr) ${}^{*}$ Corresponding author}
	%\thanks{ \noindent
	%   This paper is supported by my funding agencies.
	%  }
	
	\abstract{
		In this paper, we propose a dual memory structure for reinforcement learning algorithms with replay memory. The dual memory consists of a main memory that stores various data and a cache memory that manages the data and trains the reinforcement learning agent efficiently. Experimental results show that the dual memory structure achieves higher training and test scores than the conventional single memory structure in three selected environments of OpenAI Gym. This implies that the dual memory structure enables better and more efficient training than the single memory structure. 
	}
	
	
	\maketitle
	
	%-----------------------------------------------------------------------
	
	\section{Introduction}
	Replay memory plays an important role in stable learning and fast convergence of deep reinforcement learning algorithms \cite{ref:1} that are methods of approximating a value or a policy function using deep neural networks \cite{ref:2}. The study of replay memory in reinforcement learning started from \cite{ref:3} and played a major role in training reinforcement learning agents to play Atari 2600 games with a Deep Q-Network (DQN) \cite{ref:4}. In addition, replay memory is  used in other  off-policy reinforcement learning algorithms such as DDPG \cite{ref:5} and ACER \cite{ref:6}. In  \cite{ref:7}, after analyzing the importance of the data in the replay memory, a probability distribution is assigned to enable efficient learning through prioritization based on the importance. On the other hand, \cite{ref:8} proposed a method for stochastically eliminating data based on the importance in replay memory. However, each of the  priority-based learning method and the memory management method greatly increases the cost of computing the importance of all data as the memory capacity increases. In general, this problem may occur when deep reinforcement learning algorithms with replay memory set the memory size very large.
	
	To handle this problem, we propose a new memory structure for efficient use of replay memory in deep reinforcement learning. The proposed memory structure consists of a main memory and a cache memory. The main memory is created to store various data, and the cache memory is created to efficiently manage the data and train the agent based on the importance of  data. We compare the training performance of the proposed structure with that of the conventional single memory structure in the OpenAI Gym \cite{ref:9} environment, and verify the effect of the proposed structure.
	
	\section{Proposed Structure}
	\subsection{Dual Memory Structure}
\begin{figure}[!t]
		\centering
		\includegraphics[width=0.92\linewidth]{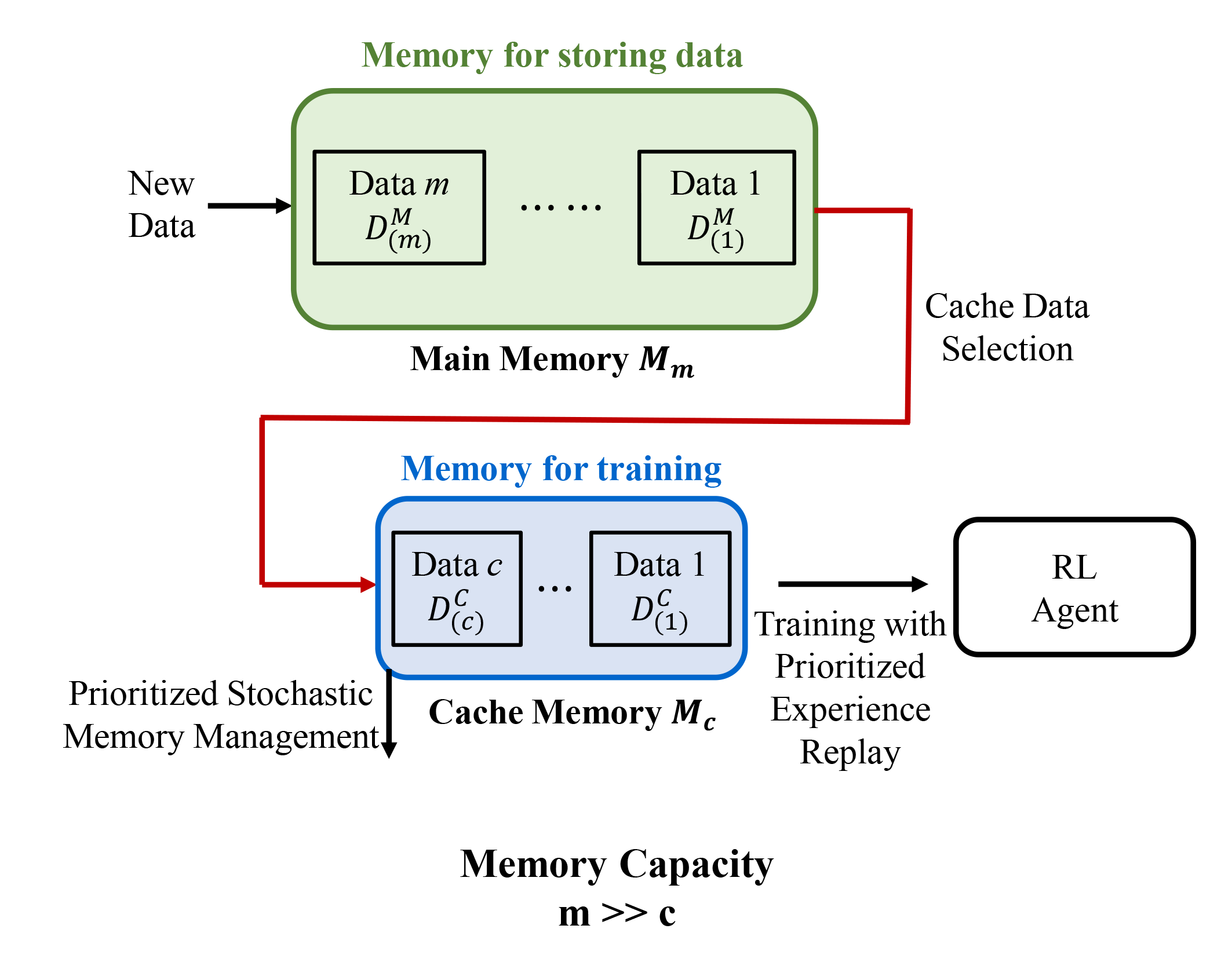}
		\caption{\label{fig:1}Proposed dual memory structure.}
	\end{figure}

	The proposed memory structure is divided into two parts. One part has a large capacity and mainly stores various data, which is called  main memory in this paper. The other  has a much smaller capacity relative to the main memory and is used to efficiently manage the data and train the agent. This memory shall be simply called cache memory. As shown in Fig. \ref{fig:1}, this memory structure first stores the new data obtained by interacting with the environment in the main memory, and copies a certain portion of the main memory to the cache memory for learning. The reason why the capacity of the cache memory is made smaller is to reduce the cost of computing the importance of each data, thereby efficiently performing both prioritized data removal and prioritized training based on the importance. So, we remove the cache memory data using the Prioritized Stochastic Memory Management (PSMM) method \cite{ref:8} and sample the training data using the Prioritized Experience Replay (PER) method \cite{ref:7}. If we use a single memory that is not separated by their role, the larger the size of memory, the higher the computation load. On the other hand, if the importance of data is calculated only in a cache memory with a small capacity, much faster calculations can be performed
	
	\subsection{Cache Data Selection}
	\begin{figure}[!t]
		\centering
		\includegraphics[width=\linewidth]{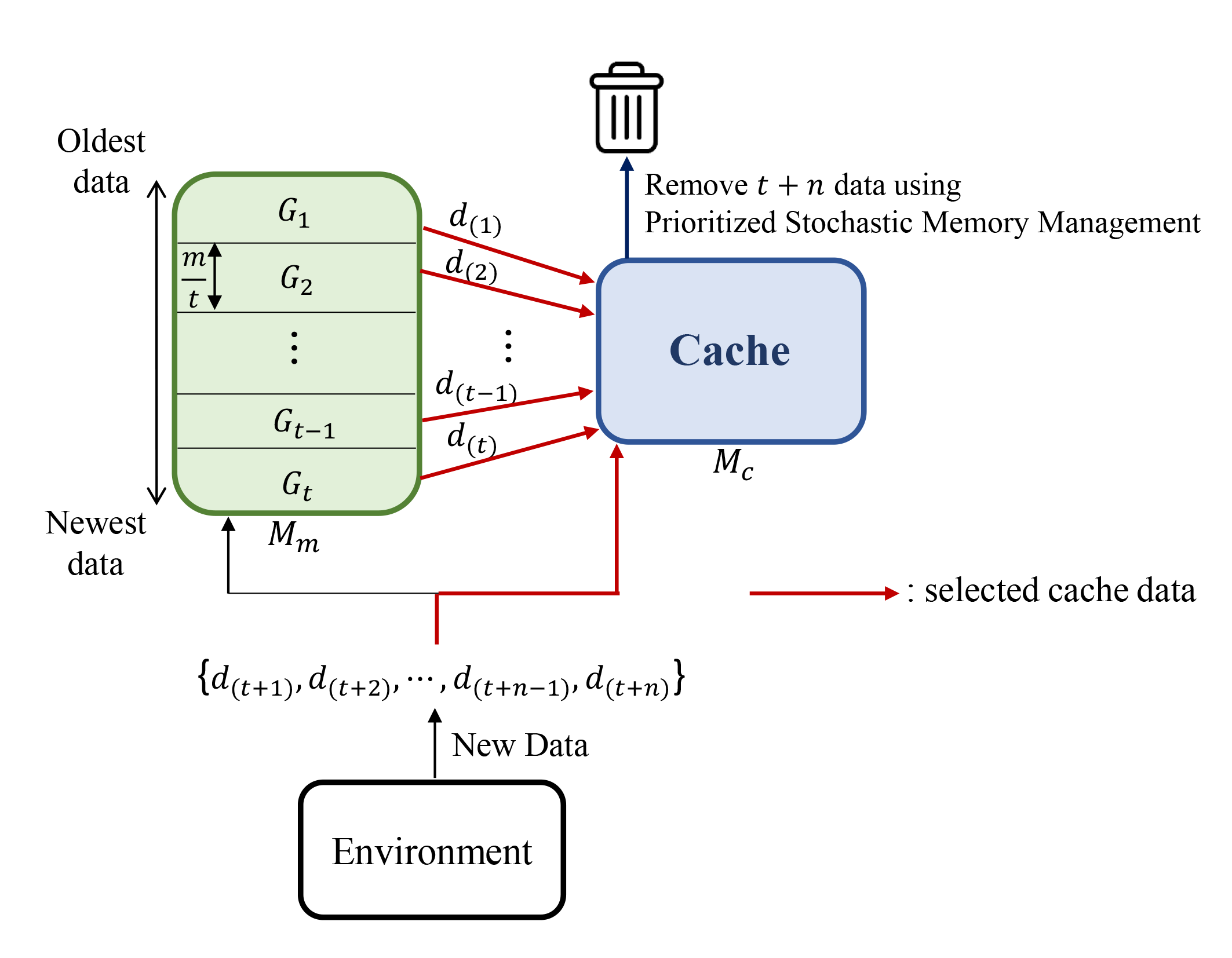}
		\caption{\label{fig:2}Cache data selection method.}
	\end{figure}
	
	\begin{table*}[!b]
		\caption{Selected cache data.}
		\centering
		\begin{tabular}{|c|c|c|c|}
			\hline
			Acquisition method & Source & \# of data & Dataset\\\hhline{|====|}
			
			Randomly sample from each subset $G$ & Main Memory & $t$ & $\left\{d_{(1)},d_{(2)},\cdots,d_{(t-1)},d_{(t)}\right\}$ \\ 
			\hline
			Get a new data by interacting with the environment & Environment & $n$ & $\left\{d_{(t+1)},d_{(t+2)},\cdots,d_{(t+n-1)},d_{(t+n)}\right\}$ \\
			\hline
		\end{tabular}
		\label{tab:1}
	\end{table*}
	
	A method for selecting data to be transferred from the main memory  to the cache memory is as follows. First, it is assumed that data in the main memory with the capacity $m$ is arranged in the order of stored time. That is, if the $i$-th data in the main memory is denoted by $D_{(i)}$, the dataset $M_m$ of the main memory can be expressed as
	\begin{equation}
	M_m=\left\{D_{(1)},D_{(2)},\cdots,D_{(m-1)},D_{(m)}\right\}.
	\label{eq:1}
	\end{equation}
	In Eq. (\ref{eq:1}), the smaller the index of of the data $D$, the older the data; for example, $D_{(1)}$ is the oldest data in $M_m$.
	Next, the entire main memory is divided into $t$ region according to stored time. In other words, we obtain $t$ disjoint subsets from $M_m$ where each subset has $\frac{m}{t}$ elements, where $m$ is assumed to be divisible by $t$. Let $G_j$ denote the subset $j$ with $j=1,\cdots,t$. Then, the first subset $G_1$ and $t$-th subset $G_t$ can be expressed as
	\begin{align*}
	&G_1=\left\{D_{(1)},D_{(2)},\cdots,D_{(\frac{m}{t}-1)},D_{(\frac{m}{t})}\right\},\\
	&G_t=\left\{D_{(\frac{(t-1)m}{t}+1)},D_{(\frac{(t-1)m}{t}+2)},\cdots,D_{(m-1)},D_{(m)}\right\}.
	\end{align*}
	So, the $j$-th subset $G_j$ can be expressed as
	\begin{align*}
	G_j=\left\{D_{(\frac{(j-1)m}{t}+1)},D_{(\frac{(j-1)m}{t}+2)},\cdots, D_{(\frac{jm}{t})}\right\}.
	\end{align*}
	In fact, the number of the divided region of the entire main memory is equal to the number of data copied from main memory to cache memory. After dividing the entire dataset with $t$ disjoint subsets as above, we randomly sample the data one by one from each subset. By doing so, we obtain a sampled dataset $\left\{d_{(1)},d_{(2)},\cdots,d_{(t-1)},d_{(t)}\right\}$ from the main memory where $d_{(j)}$ is a randomly sampled data from $G_j$.
	The reason why we divide the main memory into some region according to time and sample the data from each of the region is to obtain data in various time intervals.
	
	In addition, new data from the environment is also copied to the cache memory. If a reinforcement learning training process is not executed at every time step but  at every $n$ time steps, we copy  to the cache memory  $n$ new data generated between training steps. In Fig. \ref{fig:2}, new data from the environment is denoted by $\left\{d_{(t+1)},d_{(t+2)},\cdots,d_{(t+n-1)},d_{(t+n)}\right\}$.
	
	In summary, if we train the agent at every $n$ time steps, $t$ data from the main memory and $n$ new data from the environment are obtained and a total of $t+n$ data are copied to the cache memory at every training step; refer to Table \ref{tab:1} for summary of the selected cache data. However, if the cache memory is already full, we remove $t+n$ data from the cache memory using the Prioritized Stochastic Memory Management method \cite{ref:8} just before we copy the selected cache data to the cache memory. 
	
	\section{Experiments}
	\subsection{Experimental Setup}
	In order to compare the performance of the proposed dual memory structure with that of the single memory structure, we train reinforcement learning agents for three difference cases: \circled{1} Single memory structure with PER, \circled{2} Single memory structure with PSMM, and \circled{3} Dual memory structure with PER and PSMM. For the single memory of \circled{1} and \circled{2}, the memory is set to store 10,000 reinforcement learning data units, and for the dual memory  of \circled{3} the main memory stores 8,000 data units and the cache memory stores 2,000 units of data. 
	
	The deep reinforcement learning algorithm used for the experiment is DQN \cite{ref:4}, and the importance of the data is determined by the absolute value of the TD error; refer to \cite{ref:7} for more detail on the importance of data in DQN. In fact,  the PSMM method in \cite{ref:8} was proposed for the Actor-Critic method \cite{ref:10}, \cite{ref:11}, so the Return as well as the TD error affected the importance of the data. In these experiments, however, we consider only the TD error as the importance of the data. Each agent is trained in Assault-v0, SpaceInvaders-v0, and KungFuMaster-v0 of OpenAI Gym \cite{ref:9} using the high-quality implemented algorithm provided by OpenAI Baseline \cite{ref:12} for a total of 1 million steps. 
	\begin{figure*}[!t]
		\centering
		\begin{subfigure}{0.98\columnwidth}
			\centering
			\includegraphics[width=\linewidth]{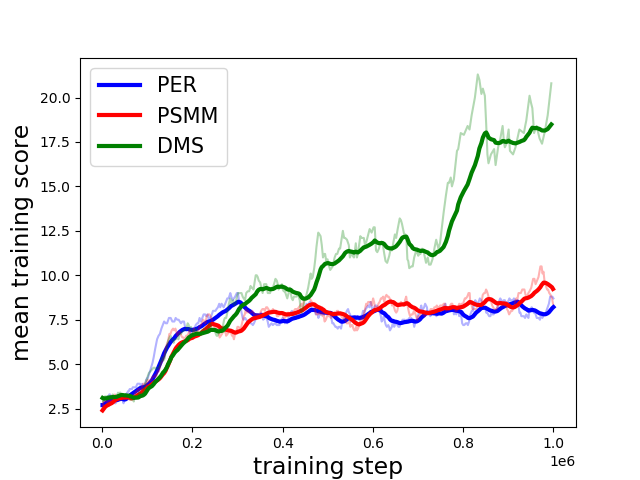}
			\caption{mean training score for past 100 consecutive episodes}
		\end{subfigure}
		\begin{subfigure}{0.98\columnwidth}
			\centering
			\includegraphics[width=\linewidth]{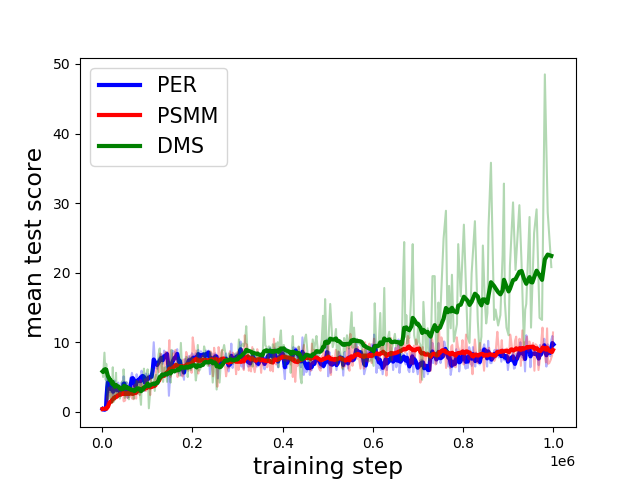}
			\caption{mean test score for 10 test episodes}
		\end{subfigure}
		\caption{The result on Assault-v0.}
		\label{fig:3}
	\end{figure*}
	
	\begin{figure*}[!t]
		\centering
		\begin{subfigure}{0.98\columnwidth}
			\centering
			\includegraphics[width=\linewidth]{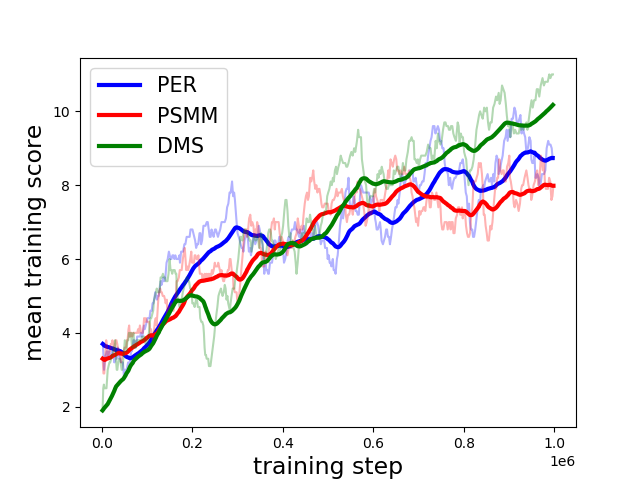}
			\caption{mean training score for past 100 consecutive episodes}
		\end{subfigure}
		\begin{subfigure}{0.98\columnwidth}
			\centering
			\includegraphics[width=\linewidth]{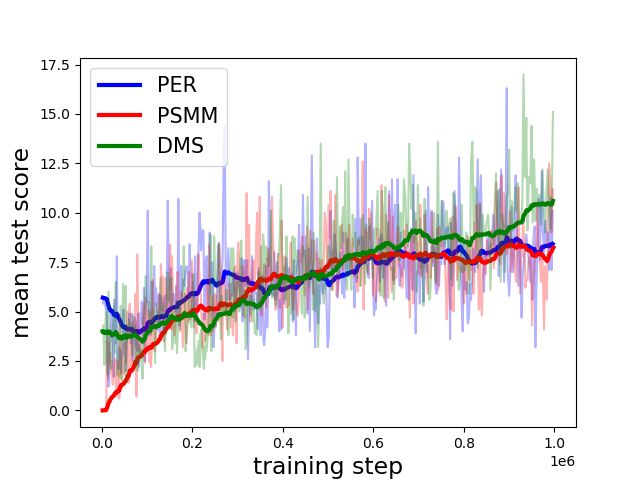}
			\caption{mean test score for 10 test episodes}
		\end{subfigure}
		\caption{The result on SpaceInvaders-v0.}
		\label{fig:4}
	\end{figure*}
	
	\begin{figure*}[!t]
		\centering
		\begin{subfigure}{0.98\columnwidth}
			\centering
			\includegraphics[width=\linewidth]{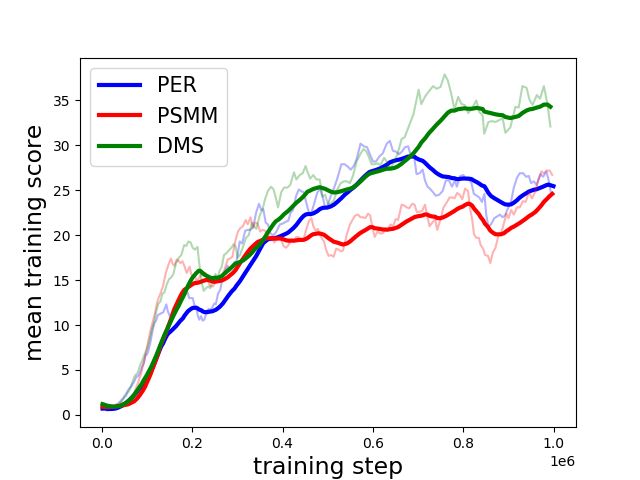}
			\caption{mean training score for past 100 consecutive episodes}
		\end{subfigure}
		\begin{subfigure}{0.98\columnwidth}
			\centering
			\includegraphics[width=\linewidth]{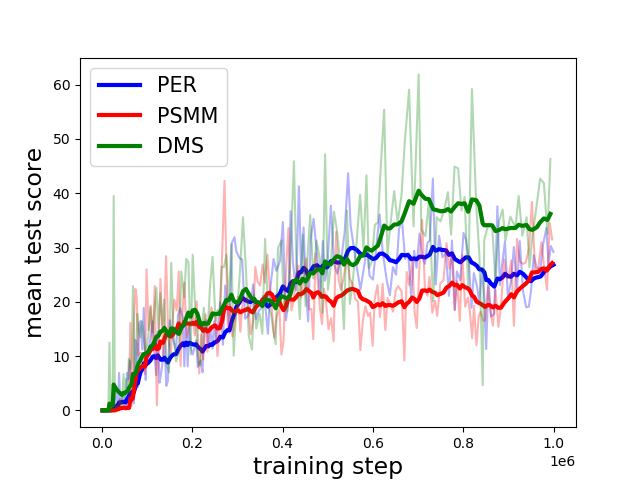}
			\caption{mean test score for 10 test episodes}
		\end{subfigure}
		\caption{The result on KungFuMaster-v0.}
		\label{fig:5}
	\end{figure*}
	
	\subsection{Results}
	Figures \ref{fig:3}$-$\ref{fig:5} show the mean training score for past 100 consecutive episodes and the mean test score for 10 test episodes during the training process in Assault-v0, SpaceInvaders-v0, and KungFuMaster-v0. Here, the blue line labeled PER indicates the result for the Single Memory Structure with PER, the red line labeled PSMM indicates the result for the Single Memory Structure with PSMM, and the green line labeled DMS indicates the Dual Memory Structure with PER and PSMM. In Figs. \ref{fig:3}$-$\ref{fig:5}, we can see that the DMS method has the highest mean training and test scores in all the three environments. In particular, in Fig. \ref{fig:3}, which is the result on Assault-v0, the mean test score of the DMS method is about five times higher than those of the two single memory methods.
	
	From the above experimental results, it can be seen that the proposed dual memory structure designed to efficiently use both PER and PSMM shows higher training performance than the single memory structure with either PER or PSMM. This means that the cache memory with a small capacity can efficiently use the PER method and the PSMM method to enhance the training performance of the reinforcement learning agent, and the main memory plays a role of maintaining diversity of data which may be lacking in the cache memory.
	
	\section{Conclusion}
	In this paper, we have proposed a dual memory structure to improve the performance of the reinforcement learning algorithm using replay memory. The memory structure is divided into a large-capacity main memory part for storing various data and a small-capacity cache memory part for efficient training. This research is an improvement on the previous study on the efficient usage of replay memory in reinforcement learning with a single memory. The experimental results show that the proposed dual memory structure with the Prioritized Experience Replay (PER) method and the Prioritized Stochastic Memory Management (PSMM) method achieves higher training and test scores than a single memory structure. This implies that the memory structure divided into two parts enables better and more efficient training than the conventional single memory structure. However,  in this paper the dual memory structure has been applied only to DQN that is an algorithm for learning the optimal discrete action. In future work, we will study the possibility of generalization of this memory structure to the continuous action case, by applying it to  DDPG and ACER that are reinforcement learning algorithms for continuous action spaces.
	
	\section*{Acknowledgement}
	This research has been in part supported by the ICT R\&D program of MSIP/IITP [2016-0-00563, Research on Adaptive Machine Learning Technology Development for Intelligent Autonomous Digital Companion].
	
	\bibliography{ICCAS2019} 
	\bibliographystyle{ieeetr}
	
\end{document}